\documentclass[10pt, conference]{IEEEtran}

\usepackage{amsmath,amssymb,amsfonts}
\usepackage{graphicx}
\usepackage{pifont}
\usepackage{float}

\usepackage{amsthm}
\usepackage{xcolor}
\usepackage[figuresright]{rotating}

\usepackage{graphicx}
\graphicspath{{/}{fig/}}

\usepackage{array}
\usepackage{textcomp}
\usepackage{xcolor}
\usepackage{multirow}
\usepackage{booktabs}
\usepackage{enumerate}

\usepackage{mathtools}
\usepackage{breqn}
\usepackage{float}

\usepackage{pgfplots}
\pgfplotsset{compat=1.7}
\usepgfplotslibrary{groupplots}

\usepackage{caption}
\usepackage{subcaption}

\usepackage{multirow,tabularx}
\usepackage{hyperref}
\usepackage{flushend}
\usepackage{algorithmic}
\usepackage[vlined, ruled, shortend]{algorithm2e}

\usepackage{pgfplots}
\pgfplotsset{compat=1.7}
\usepgfplotslibrary{groupplots}

\usepackage[font=small]{caption}
\usepackage{subcaption}

\usepackage{multirow,tabularx}
\usepackage{hyperref}
\usepackage{flushend}
\usepackage{algorithmic}
\usepackage{soul}

\usepackage[vlined, ruled, shortend]{algorithm2e}

\usepackage{pgfmath}
\usetikzlibrary{decorations.text, arrows.meta,calc,shadows.blur,shadings, spy}

\usepackage{listings}
\newlength\figureheight
\newlength\figurewidth
\SetAlCapNameFnt{\footnotesize}
\SetAlCapFnt{\footnotesize}

\title{
    A Customizable Conflict Resolution and Attribute-Based Access Control Framework for Multi-Robot Systems
}

\author{
    \IEEEauthorblockN{
        Salma Salimi$^{*,}$\IEEEauthorrefmark{2},
        Farhad Keramat$^{*}$\IEEEauthorrefmark{2},
        Tomi Westerlund\IEEEauthorrefmark{2},
        Jorge Pe\~na Queralta\IEEEauthorrefmark{2}
    }
    \IEEEauthorblockA{
        \normalsize
        $^{*}$ These authors contributed equally. \\ \IEEEauthorrefmark{2}\href{https://tiers.utu.fi}{Turku Intelligent Embedded and Robotic Systems (TIERS) Lab, University of Turku, Finland}.\\
        Emails: \textsuperscript{1}\{salmas, fakera, tovewe, jopequ\}@utu.fi\\[+6pt]
    }
}

\begin{document}

\maketitle
\thispagestyle{empty}
\pagestyle{empty}



\begin{abstract}\label{sec:abstract}%
    As multi-robot systems continue to advance and become integral to various applications, managing conflicts and ensuring secure access control are critical challenges that need to be addressed. Access control is essential in multi-robot systems to ensure secure and authorized interactions among robots, protect sensitive data, and prevent unauthorized access to resources. This paper presents a novel framework for customizable conflict resolution and attribute-based access control in multi-robot systems for ROS\,2 leveraging the Hyperledger Fabric blockchain. We introduce an attribute-based access control (ABAC) Fabric-ROS\,2 bridge to enable secure communication and control between users and robots. By defining conflict resolution policies based on task priorities, robot capabilities, and user-defined constraints, our framework offers a flexible way to resolve conflicts. Additionally, it incorporates attribute-based access control, granting access rights based on user and robot attributes. ABAC offers a modular approach to control access compared to existing access control approaches in ROS\,2, such as SROS2. Through this framework, multi-robot systems can be managed efficiently, securely, and adaptably, ensuring controlled access to resources and managing conflicts. 
    Our experimental evaluation shows that our framework marginally improves latency and throughput over exiting Fabric and ROS\,2 integration solutions. At higher network load, it is the only solution to operate reliably without a diverging transaction commitment latency. We also demonstrate how conflicts arising from simultaneous control or a robot by two users are resolved in real-time and motion distortion is effectively eliminated.
\end{abstract}


\IEEEpeerreviewmaketitle


\section{Introduction}\label{sec:intro}
\begin{figure}
    \centering
    \includegraphics[width=0.49\textwidth]{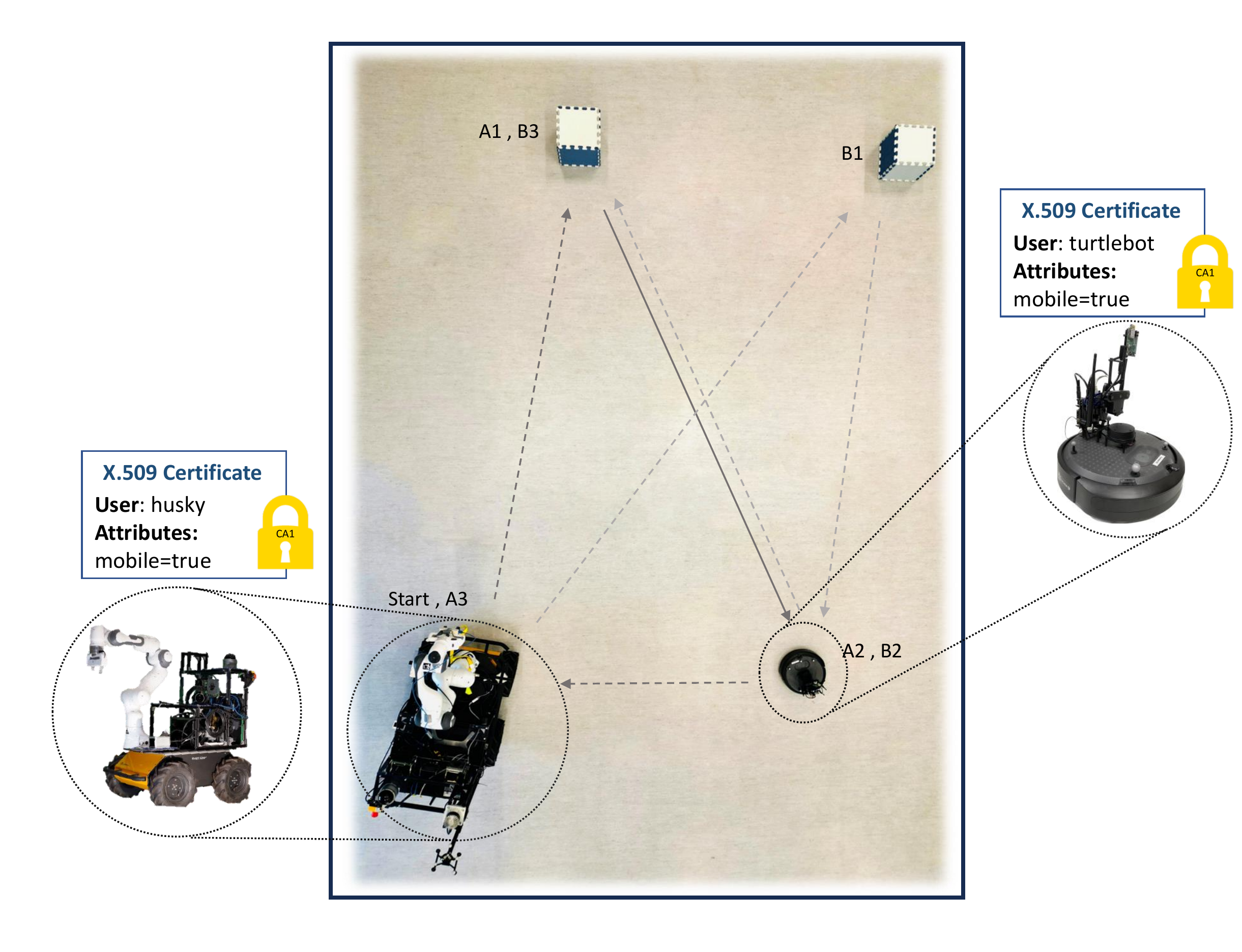}
    \caption{An attribute-based access control framework presented in this paper is used to solve conflicts between two robots. User-defined attributes are used to grant or deny access to resources, and the conflict resolution mechanism ensures that shared resources are accessible and fair. The paths illustrate simple tasks assigned by users to the robots, highlighting its practical application in multi-robot systems.}
    \label{fig:experiment}
\end{figure}


In an increasingly interconnected world, robots are being deployed at a bigger scale. This offers the potential for introducing distributed networked systems approaches and technologies~\cite{simoens2018internet, zhang2022distributed}. Indeed, blockchain technologies have been already incorporated into robotic systems to address security and interoperability challenges, among others~\cite{ferrer2018blockchain, strobel2018managing, ferrer2021following}. However, there is some skepticism in the robotics community about the scalability of these systems and whether they are suitable for real-world use cases~\cite{queralta2023blockchain}.

Recent research has started to provide more insight into the practical applications of blockchain in the context of robotics. For example, next-generation distributed ledger technologies (DLTs), such as IOTA, go beyond linear blockchains and enable more efficient and scalable consensus mechanisms adaptable to real-world connectivity conditions~\cite{keramat2022partition}. Additionally, permissioned blockchains have demonstrated more realistic use cases by allowing for greater control and customization of access and governance. 

Most of the current development in blockchain and robotics utilizes the open and permissionless Ethereum network. The Ethereum Virtual Machine enables the use of smart contracts that have led to the creation of various applications including secure federated learning~\cite{ferrer2018robochain} and reliable vehicular networks~\cite{xianjia2021flsurvey}. Nonetheless, Ethereum is encountering difficulties with scalability and meeting the requirements of industrial robotic applications, despite implementing proof-of-stake consensus measures~\cite{queralta2021blockchain}.


Hyperledger Fabric, along with other permissioned blockchain solutions, inherently possesses robust identity management and data access control mechanisms, making them highly suitable for industrial use cases and private network configurations. On the other hand, a key element of the Hyperledger Fabric blockchain is its ability to manage identity via certificate authorities, control data access policies, and create private data channels, which can facilitate its adoption within the robotics community. In our previous works, we have demonstrated the potential of using Fabric-ROS\,2 interfaces for various applications such as data recording~\cite{salimi2022towards}, multi-robot collaboration~\cite{salimi2022secure}, cooperative decision-making processes~\cite{torrico2022UWBRole}, and also real-time two-way data transmission~\cite{lei2023event}. 
However, the potential benefits of leveraging Fabric as a permissioned blockchain for access control and protection of private data have yet to be fully explored in existing literature. Previous studies integrating ROS\,2 and Fabric have been unable to effectively restrict unauthorized access to transferred data by nodes or robots within the blockchain network.
To address this gap, this paper introduces a framework that employs Attribute-Based Access Control (ABAC) to grant access solely based on the attributes provided by the intended recipient.

Access control is a security measure that either permits or denies access rights to a user based on their perceived level of risk~\cite{zhang2020attribute, riabi2019blockchain}. It comprises three crucial components: authentication, authorization, and auditing. An authentication process verifies a user's identity, while an authorization process determines whether or not they are authorized to perform certain operations on an object. Auditing, on the other hand, facilitates post-activity analysis of the system. Despite the importance of all three components in system security, authorization is the most crucial because it enforces access policies. However, if access control systems are not properly implemented, they can result in significant privacy and financial losses for individuals and organizations~\cite{pinno2017controlchain}.

Attribute-based access control (ABAC) is an effective decentralized access control model due to its scalability, dynamic nature, and versatility. As part of the ABAC process, the attributes of operations, entries, and their contexts are evaluated to determine access between subjects and objects~\cite{liu2020fabric}. Additionally, ABAC can provide granular and sophisticated access control, making it highly expressive.

Our implementation of ABAC on the blockchain utilizes smart contracts. With this approach, the access control can be fine-grained and expressive. ABAC is a highly suitable decentralized access control model for IoT scenarios due to its scalability, flexibility, and dynamic nature. In order to transform permission management into attribute management, ABAC extracts and combines attributes from the subject (user), object (resource), permissions, and environment. Access management systems resulting from this process are highly sophisticated and dynamic~\cite{liu2020fabric}.

The structure of this paper is as follows: Section II delves into the contextual background and prior research concerning the utilization of blockchain technology for robotics, along with the integration of Hyperledger Fabric and ROS\,2. Section III introduces the methodology utilized in the study, while Section IV presents the experimental results. The scalability of the approach is discussed in Section V and finally, the paper concludes in Section VI.

\section{Background}
\label{sec:related}

Through this section, we review the background concepts and relevant literature in permissioned blockchain networks, ABAC applications, security in ROS\,2, and Fabric-ROS\,2 integration.



When the Robot Operating System (ROS) was first developed, security was not a primary concern. However, as it became more widely used in various applications and government programs, the need for security measures became more apparent. Initial efforts to improve security in ROS included several research papers, such as \cite{dieber2016application, white2016sros, lera2016ciberseguridad}, which were published in 2016. Currently, these early security efforts are no longer being actively maintained and the focus of the ROS community has shifted to improving security in ROS\,2.
A usable security approach to robotics has been argued in~\cite{mayoral2022sros2} as the best method to remain secure in robotics and SROS\,2 is introduced for adding security features to ROS\,2, as security in robotics is a process that requires continuous evaluation on a periodic basis, not a product~\cite{mayoral2022robot, zhu2021cybersecurity, mayoral2020alurity}.
In other words, SROS\,2 is an updated version of ROS\,2 that includes security enhancements, such as socket transport security and access control mechanism~\cite{breiling2017secure, white2016sros}. Publishers/subscribers defined in a policy file are only allowed to publish/subscribe to one topic under the provided access control model~\cite{white2018procedurally}. 
As an additional means of improving security, blockchain technology can be integrated with ROS\,2.



Bitcoin and Ethereum are popular blockchains that are permissionless, meaning anyone can join and view all the recorded data without permission. However, this technology cannot be used in scenarios where organizations want to exchange data without making it publicly available or sharing it with other parties on the same blockchain. As a permissioned blockchain, Hyperledger is only trusted by a subset of peers, and nodes are not privileged equally~\cite{androulaki2018hyperledger}. As a result, the blockchain technology can be incorporated into more controlled environments. The permissioned blockchain protects data privacy by enforcing access control mechanisms and restricting transactions to only authorized participants~\cite{wang2021private}. They also typically use deterministic consensus mechanisms for fast consensus among authenticated users, making them suitable for enterprise applications which require deterministic processing of high volumes of transactions~\cite{xu2021latency}.

The Hyperledger Fabric blockchain platform was created with the intention of being used by businesses and has certain features that are useful for distributed robotic systems. These include the ability to identify participants and manage their identities, as well as generate certificates. The networks are permissioned, which means that there is built-in security for data. Additionally, the Hyperledger Fabric can handle high transaction throughput and confirm transactions in real-time with configurable low-latency transaction confirmation. Finally, ROS\,2 topics and the associated pub/sub system in the underlying DDS communication can seamlessly be extended with partitioning of data channels and privacy and confidentiality of transactions.


It is evident that DLTs can bring valuable features to robotic systems, such as the ability to establish trust within a decentralized network of agents. While DLTs are not typically found in common robotic middlewares like ROS\,2~\cite{macenski2022robot, mayoral2022sros2}, they can be integrated with other platforms to build secure and reliable robotic systems. For instance, the integration of ROS\,2 with Hyperledger Fabric can be used to ensure data integrity and traceability in autonomous robot systems, as well as secure communication and collaboration between multiple robots and other components. Moreover, in supply chain management applications, Hyperledger Fabric can provide an immutable record of transaction history, ensuring transparency and accountability across the entire supply chain. By leveraging the strengths of both technologies, this integration can create more sophisticated and resilient robotic systems.

The framework presented in~\cite{salimi2022towards} describes the use of Hyperledger Fabric to control robots, collect and process data, and integrate with ROS\,2. Furthermore, using dynamic UWB role allocation algorithms, this framework in~\cite{torrico2022UWBRole} is applied to distribute decision-making in a multi-robot system.
The framework provides enhanced identity and data access management, making it secure and reliable. \cite{salimi2022secure} also uses Fabric and ROS\,2 on a multi-robot inventory management task to drive interaction and role allocation between the robots. The smart contracts in the blockchain are used for high-level mission control, such as instructing robots on predefined paths or returning to a designated position for charging. However, the methods presented in~\cite{mokhtar2019blockchain} for multi-robot path planning are tailored to a specific problem and algorithm, limiting potential applicability or integration to other domains.

Although Fabric has proven to be beneficial and have built-in properties for industrial systems, its suitability as a platform for remote robot control or teleoperation had not been explored before. With the help of a novel event-driven integration approach presented in~\cite{lei2023event}, Fabric blockchain can transmit data between ROS\,2 systems in the hundreds of milliseconds, similar to commonly available mobile networks. There are several applications and use cases that could be derived from this discovery.

SROS\,2 incorporates built-in role-based access control (RBAC), while blockchain frameworks, in contrast, do not possess this inherent access control feature. However, despite this limitation, it is possible to integrate or implement access control within a smart contract. 

Access control to data in information systems is typically either role-based, ruled-based, policy-based or attribute-based, among other approaches including discretionary access. ABAC uses the attributes of the object, subject, permission, and environment to decide on granting access to a requester or not. A request's outcome is determined by whether the necessary attributes are present in the request. Target systems can also use these attributes to define access policies to determine if a requester has sufficient privileges. ABAC separates access control and policy management effectively by defining subject and object attributes separately~\cite{zhang2020attribute, liu2020fabric}. As a result, an individual does not need to be explicitly authorized to perform operations on objects. ABAC uses standardized subject and object attribute definitions, which is consistent across organizations~\cite{figueroa2019attribute}.

Access control frameworks are widely studied in robotics, for instance, to address the complexity of data characteristics and access patterns associated with connected and autonomous vehicles (CAVs), \cite{zhang2020ac4av} presents a three-layer access control framework. Its prototype implementation of a data access control framework demonstrates its effectiveness while providing extensibility for third-party developers to implement their own access control models.

Another critical issue that arises alongside access control in robotics is the challenge of conflict resolution. As robotics proliferate across various industries, conflicts between humans and robots can arise, particularly when multiple users attempt to access the same robot at the same time~\cite{babel2022human,babel2021development}. Resources, including the robot itself, are limited, which leads to such conflicts so for robotic systems to operate smoothly and effectively, conflict resolution mechanisms are essential. In order to avoid disruptions and optimize the efficiency of human-robot interaction, these mechanisms establish protocols and procedures to manage conflicts and prioritize access to robots. The integration of robots into industries can be made possible through conflict resolution strategies, which will enhance human-robot collaboration.

Our contribution to the literature involves integrating Hyperledger Fabric with an attribute-based access control (ABAC) smart contract within the ROS\,2 framework, enhancing the role-based access control (RBAC) capability of SROS\,2. Unlike RBAC, which requires policy reconfiguration when adding new users after the initial setup, our approach based on attribute-based access controls defines policies based on user attributes rather than identities. This eliminates the need for system modifications when incorporating new users. Additionally, our integration provides conflict resolution mechanisms, ensuring data integrity and consistency. By combining these advancements, our solution strengthens the security and reliability of SROS\,2, making it well-suited for a variety of robotic applications.

\section{Methodology}

\begin{figure*}[t]
    \centering
    \includegraphics[width=1.0\textwidth]{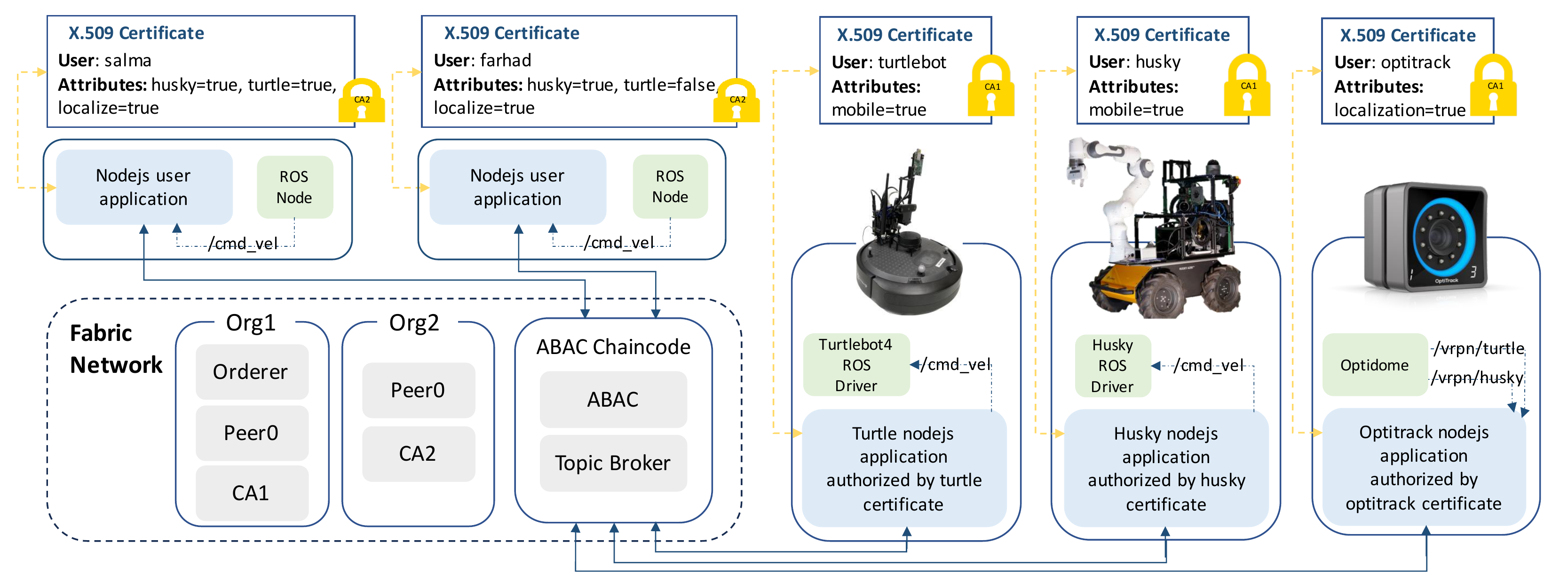}
    \caption{The system architecture for the intra-organizational collaboration problem defined with attributed-based access control and conflict resolution. X.509 certificates for each entity in the system are depicted with the corresponding attributes and the CA that has signed the certificate. Each entity including robots and users runs an application that is responsible to communicate with the Fabric network. The subscribers and publishers in each entity are demonstrated with dashed arrows with the name of the topic on the arrow.}
    \label{fig:architecture}
\end{figure*}

In this section, we define the problem, and our proposed solution to it. We will also discuss the equipment and techniques employed in our study. Specifically, we will outline the system's network configuration and the hardware configuration of the robot models used.

\subsubsection{\textbf{Problem formulation}}

Consider the intra-organizational collaboration between organizations $O_1$ and $O_2$. Organization $O_2$ provides multiple robots $\{\mathcal{R}_i\}_{i\in1\dots N}$ as services for $O_1$. Each robot $\mathcal{R}_i$ offers different services $\mathcal{S}_{\mathcal{R}_i}$ that $\mathcal{S}_{\mathcal{R}_i}$ and $\mathcal{S}_{\mathcal{R}_j}$ are not necessarily the same for $\mathcal{R}_i$ and $\mathcal{R}_j$. We define each service as the right to publish/subscribe on a ROS\,2 topic. Two types of operation modes are defined for each service. In open operation mode, any user that has access to the service can publish on the corresponding topic. Alternatively, in the exclusive operation mode, only one of the users with access can publish on the topic. Other than these, other operation modes can be considered that we do not implement them in this work. For example at the same time, $k_1$ users can publish and $k_2$ users subscribe to the topic. Also, it is possible to prioritize users requesting a service from a robot based on other metrics.

After the initial setup, continuously new robots can join $O_2$ or a robot can get out of service. $O_1$ has several users $\mathcal{U}$ who will benefit from $O_2$ robots. $O_1$ organization wants to give access to each user only the required services from robots. Each user $u_i \in \mathcal{U}$ will get access to a subset of services. New users can join the organization $O_1$ or their access to the robots can be revoked at any time.

The problem of an access control system with conflict resolution is highly customizable. The benefit of our proposed method using Hyperledger Fabric chaincode is the possibility to implement such a highly customizable setup in contrast to the access control offered in SROS\,2. Also, our proposed method is completely modular and does not affect the ROS\,2 nodes implementation while implementing such a customizable conflict resolution platform with SROS\,2 requires major changes on ROS\,2 layer. For example to give access only to one of the authorized users.

\subsubsection{\textbf{Task}}

We define a task based on the general problem defined previously. In this setup, $O_2$ has three robots a Turtlebot4, a Husky, and an OptiTrack camera giving services to $O_1$ users \textit{Salma} and \textit{Farhad}. Here we consider that Turtlebot4 and Husky are mobile robots each of them only having one service. This service is the ability to publish on their $/cmd\_vel$ topic to control them. $O_1$ wants to give $Salma$ access to control both of the robots. On the other hand, $Farhad$ should only have access to Turtlebot4. The operation mode for this service is exclusive since robots can navigate to only one point at a time. OptiTrack camera also has a service publishing the location of robots to users. All users with access to the service can subscribe to the location values published by the OptiTrack camera. The user $u_i$ gives a set of points $\mathcal{P}_j^i$ to a robot $\mathcal{R}_j$ to traverse through them in order.
So a \textit{task} is defined as $T=(u_i, \mathcal{R}_j, \mathcal{P}_j^i)$. Here the service is omitted since in our setup each robot offers only one service. If the user $u_i$ is granted access for task $T$, then $\mathcal{R}_j$ should traverse through all points in $\mathcal{P}_j^i$.

\subsubsection{\textbf{Environment}}
The environment consists of a $8m\times9m$ square arena with a height of $5m$. Robots can navigate freely in this area and since this area is completely covered by OptiTrack cameras their position will be published at a constant rate.

\subsubsection{\textbf{Robots model}} For our experiment, we employed two different robot platforms: the Clearpath Husky with a robotic arm and the Turtlebot4 as the mobile robot. The Turtlebot4 initially had a single Raspberry Pi 4 Model B, but we later integrated a Jetson Nano into its setup to enhance its computational capabilities~\cite{yu2023loosely}. Furthermore, we also utilized the localization information obtained from the OptiTrack cameras.

Regarding the experimental site shown in~\ref{fig:experiment}, the robots navigate within an arena equipped with an Optitrack motion capture system which serves as a reliable source of ground truth data where this motion capture system accurately tracks the movements of the robots.

\subsubsection{\textbf{Proposed Configuration}}

Our proposed system configuration to address the task described above is depicted in Fig.~\ref{fig:architecture}. The system is built on top of the Hyperledger Fabric network. The main logic for access control and conflict resolution is implemented on a chaincode. This chaincode is decoupled from the ROS\,2 stack. Fabric applications that act as ROS\,2 nodes bridge the local ROS\,2 systems with the Fabric network on each robot or user terminal.

\paragraph*{Fabric Network Setup}
There are two organizations \textit{Org1} and \textit{Org2}, representing $O_1$ and $O_2$ respectively. Each organization has one peer node that is the gateway of the organization's users to the Fabric network. There is one \textit{Orderer} node in the \textit{Org1} which runs an instance of Raft~\cite{ongaro2014search} algorithm. In order to ensure secure communication between network participants, Hyperledger Fabric utilizes a Public Key Infrastructure (PKI). Participants and their public keys are identified in Hyperledger Fabric using X.509 certificates. In order to obtain these certificates, participants must first submit their identity to a Certificate Authority (CA), which is a trusted entity that verifies the identity of each participant. Each organization has its own CA. Certificates are signed by CAs using their private keys, which can be verified using their public keys. By ensuring the authenticity of both public keys and identities, X.509 certificates facilitate secure communication between network participants.

When creating X.509 certificates, CAs can assign attributes for users. For example, in Fig.~\ref{fig:architecture} you can see that user \textit{Salma} has 3 attributes, which are signed by the CA. With these attributes embedded in the certificate, we can assure that the user has a specific attribute securely.

\paragraph*{Chaincode}
The chaincode serves as the central component of the suggested model. All users must obtain authentication from a certificate authority (CA) before they can participate in the system. Through an API exposed to users and smart gateways, the chaincode primarily manages attribute-based user rights and performs user authentication for resource access. To clarify the functionality of the chaincode, we have split it into two sub-components. The main structure of the chaincode implemented in Golang is depicted in Listing~\ref{lst:chaincode}. The $Robot$ structure is defined to store each robot's information which will be used by both sub-components. Each robot is stored as an asset on the ledger. The $setup()$ method should be called by the admin of \textit{Org1} in order to create the corresponding asset for the currently available robots which in our current setup are Husky, Turtlebot4, and OptiTrack.  Other methods are part of the ABAC.

The topic broker is responsible for routing messages between users. It uses the $set()$ method from the chaincode. We use an event-driven model presented in ~\cite{lei2023event}. For every ROS\,2 message that is going to be transmitted over the Fabric network, an asset is created. The subscriber's application receives a notification upon the creation of a new asset. The topic broker is also linked to the ABAB sub-component and will check if the publisher of the topic is authorized or not. We will show in the experiments that this authorization step doesn't add extra latency to the system.

Attributed Based Access Control (ABAC) grants or denies access to services based on a set of attributes associated with the requesting user. In this sub-component, access decisions are based on policies defined in terms of attribute values. These policies can be defined by administrators of the \textit{Org1} and embedded in the chaincode. When a user requests access to a service which in this case is to publish to a topic, the ABAC system evaluates the attributes associated with the user and the service and makes a decision depending on the policies defined. This policy depends on the operation mode for example in the case of controlling a mobile robot with $/cmd\_vel$ only one user can get access.

We have implemented 3 main methods to define the policies. The $acquire()$ method described in detail in Algorithm~.\ref{alg:acquire} should be called by the user that is going to control a robot. As an input, the user provides the identity of the robot that is going to be controlled. After checking if the user has the required attribute, and if the robot has not already been acquired by another user, access will be granted. In this method, we can also implement another kind of operation mode. The user must call the $release()$ method upon finishing their task with the robot. The $authorize()$ method is the link between the ABAC sub-component and the topic broker. The topic broker leverages this method to check the authorization of any user.

\begin{lstlisting}[float,caption={Proposed Chaincode},captionpos=b,label={lst:chaincode}]
package main

type Robot struct {
	Name           string `json:"Name"`
	SubTopic       string `json:"SubTopic"`
	PubTopic       string `json:"PubTopic"`
	Operator       string `json:"owner"`
	UnderOp        bool   `json:"UnderOp"`
}

func setup(...) {...}
func acquire(...) {...}
func release(...) {...}
func GetSubmittingClientIdentity(...) {...}
func authorize(...) {...}
func set(...) {
    ...
    // Check if the publisher is authorized
    result, err = authorize(stub, args)
    if err != nil {
	return fmt.Errorf("Client not authorized!")
    }
    ...
}
\end{lstlisting}

\begin{algorithm}[t]

    \caption{Acquiring a robot}\label{alg:acquire}
    \KwIn{$ID_{user}$, $ID_{robot}$}
    \KwOut{$True$ or $False$}
    \vspace{4mm}
    $robot \gets readAsset(ID_{robot})$\;
    \If{$robot.UnderOperation$}{
        \Return{$False$}
    }
    \vspace{2mm}
    $attribute \gets getAttribute(robot)$\;
    \If{$assertAttribute(ID_{user}, attribute)$}{
        $robot.UnderOperation \gets True$\;
        $robot.Operator \gets ID_{user}$\;
        $writeAsset(ID_{robot}, robot)$\;
        \Return{$True$}
    }
    \vspace{2mm}
    \Return{$False$}
    
\end{algorithm}

\begin{figure*}[tb]
    \begin{subfigure}[t]{0.5\textwidth}
        \centering
        \setlength{\figureheight}{.5\textwidth}
        \setlength{\figurewidth}{\textwidth} 
        \scriptsize{\input{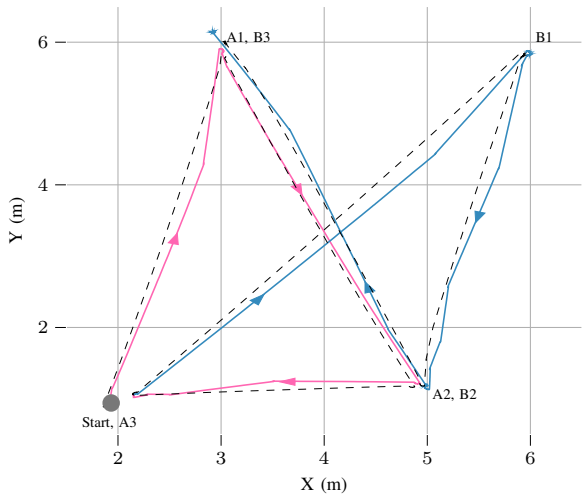}}
        \caption{Trajectory tracking using Fabric Bridge}
        \label{fig:traj_withfabric}
    \end{subfigure}
    \vspace{2em}
    \begin{subfigure}[t]{0.5\textwidth}
        \centering
        \setlength{\figureheight}{.5\textwidth}
        \setlength{\figurewidth}{\textwidth} 
        \scriptsize{\input{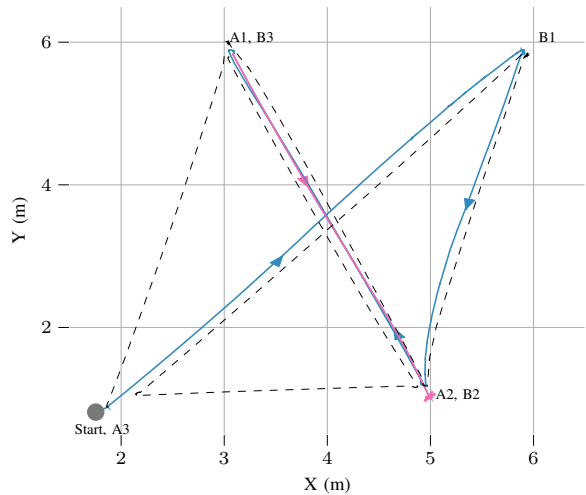}}
        \caption{Reference trajectory tracking without using Fabric bridge.}
        \label{fig:traj_nofabric}
    \end{subfigure}
    \vspace{-1em}
    \caption{This figure displays the results of real-world experiments involving a robot assigned to two tasks: traveling from Start to A3, and then from Start to B3. Subfigure (a) compares the reference trajectory with the actual trajectory of the Turtlebot4 while following the tasks with Fabric Bridge. Subfigure (b) provides the same comparison but without the use of Fabric Bridge.}
    \label{fig:trajectory}
\end{figure*}

\section{Experimental Results}

This section presents the experimental results, demonstrating how the proposed framework can be used to manage access to ROS\,2 robots through a Fabric blockchain, and highlighting the enhancements compared to prior solutions.

\begin{figure}[!ht]
    \centering
    \setlength\figureheight{0.28\textwidth}
    \setlength\figurewidth{0.49\textwidth}
    \scriptsize{\input{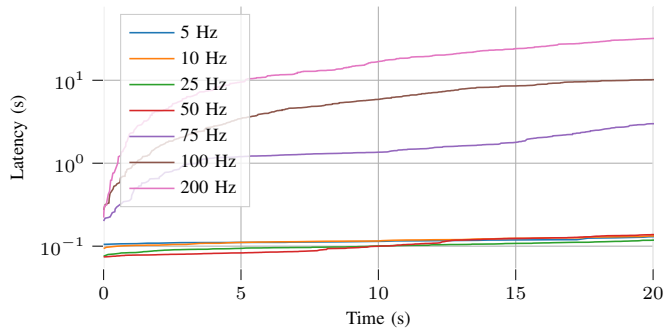}}
    \caption{This plot shows the results of a latency test conducted over Wi-Fi between a robot and a peer node, for transactions committed at various frequencies} 
    \label{fig:latency}
\end{figure}

To evaluate the scalability and performance of the system under realistic workloads, we conducted a series of tests with varying throughputs. We determined the optimal frequency for sending messages through the Fabric network to minimize latency by examining the results shown in~\ref{fig:latency}. Our experiment indicated that the optimal frequency for minimizing latency in transaction commitments was 50Hz which we have used for our experiment.
 
To verify the immediate functionality of the proposed framework, we conducted an experiment in which the Turtlebot4 in the MOCAP arena was subjected to two sets of reference trajectories shown in Figure~\ref{fig:trajectory}, Figure~\ref{fig:traj_withfabric} with and Figure~\ref{fig:traj_nofabric} without the integration of a Fabric network. Two users, both possessing Turtlebot attributes, sent commands to the robot with a Python ROS\,2 node that receives position feedback from OptiTrack cameras and sends $/cmd\_vel$ messages. Each task is to traverse through 3 waypoints in order. In the presence of the Fabric network, the robot executed the tasks sequentially, completing the first task before initiating the second one. The robot followed a specific path, moving from starting point to A1 and finally to A3. After finishing the first task it heads to B1 and finally to B3. However, in the absence of the Fabric bridge, the robot initially attempted to follow the first task, but when the second task was given, it changed its direction and started following the second task. As a result, the robot failed to maintain the correct order of the waypoints and deviated from the intended sequence.

We can see in the presence of Fabric network, because of the introduced network delay, the controller can not exactly follow the reference lines. The controller relies on the position feedback received from OptiTrack cameras which will arrive at the controller almost $300\,ms$ later~\cite{lei2023event}. However, this delay does not prevent the controller to reach all the defined waypoints. By optimizing correct linear and angular speeds according to the average network delay, we can minimize the error.

Even though the trajectories in Fig.~\ref{fig:traj_nofabric} look smoother and straight like the reference trajectory, the robot spins while traversing through waypoints since the two controllers are constantly sending $/cmd\_vel$ messages to the robot. This spinning is not visible in the trajectory plot since we do not observe the changes in the orientation of the robot.

\begin{figure}[tb]
    \begin{subfigure}[t]{0.48\textwidth}
        \centering
        \setlength{\figureheight}{0.25\textwidth}
        \setlength{\figurewidth}{0.49\textwidth} 
        \scriptsize{\input{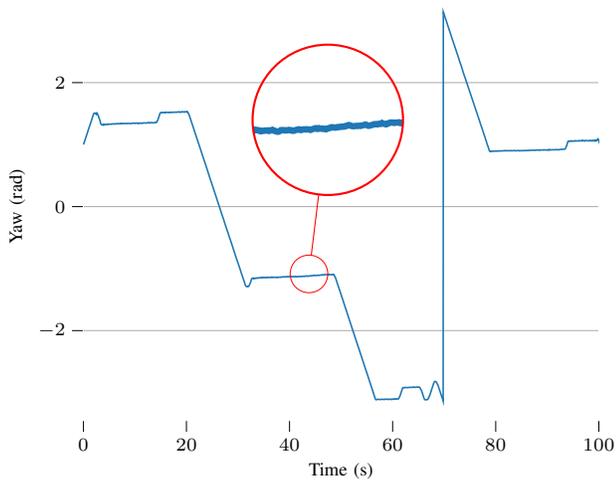}}
        \caption{Turtlebot orientation With Fabric bridge}
        \label{fig:orientation_withfabric}
    \end{subfigure}
    \vspace{1em}
    \begin{subfigure}[t]{0.48\textwidth}
        \centering
        \setlength{\figureheight}{0.25\textwidth}
        \setlength{\figurewidth}{0.49\textwidth} 
        \scriptsize{\input{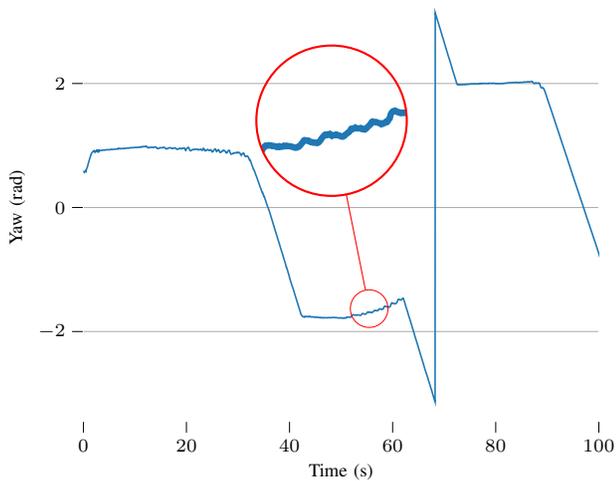}}
        \caption{Turtlebot orientation without Fabric bridge}
        \label{fig:orientation_nofabric}
    \end{subfigure}
    \vspace{-1em}
    \caption{The figure demonstrates the effects of conflict resolution on the robot's yaw when a fabric bridge is present or absent. In the absence of a fabric bridge, the robot experiences dual controller commands, resulting in noticeable vibrations. However, with the integration of a fabric bridge, these vibrations are mitigated.}
    \label{fig:ori}
\end{figure}

The comparison between Fig.~\ref{fig:orientation_withfabric} and Fig.~\ref{fig:orientation_nofabric} highlights the significance of conflict resolution in the context of integrating a Fabric bridge. When the robot executes assigned tasks sequentially with the Fabric bridge, there is no additional yaw rotation, as depicted in Fig.~\ref{fig:orientation_withfabric}. However, Fig.~\ref{fig:orientation_nofabric} showcases the detrimental consequences of conflicting control inputs. In this scenario, the robot is assigned two tasks without the presence of the Fabric bridge, resulting in erratic behavior. The figure illustrates how introducing the second task while the robot is still engaged in the first task leads to complications. The robot requires extra time to determine the correct direction to follow, thereby prolonging the task completion times. This demonstrates the critical role of conflict resolution in ensuring efficient task execution and minimizing delays.

\begin{figure}
    \centering
    \setlength\figureheight{0.28\textwidth}
    \setlength\figurewidth{0.49\textwidth}
    \scriptsize{\input{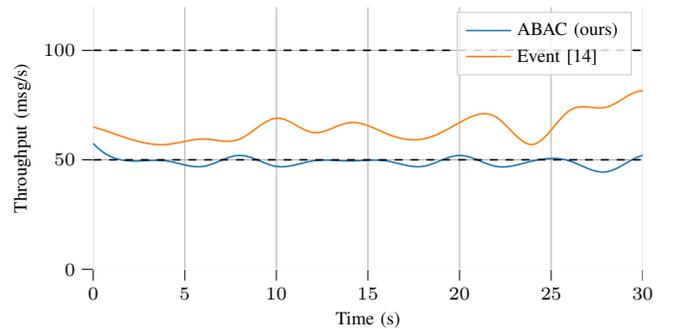}}
    \caption{This plot illustrates the results of a throughput test conducted for ABAC smart contract and event-driven smart contract. The graph shows the frequency comparison between the two} 
    \label{fig:throughput_hijack}
\end{figure}

\begin{figure}
    \centering
    \setlength\figureheight{0.28\textwidth}
    \setlength\figurewidth{0.49\textwidth}
    \scriptsize{\input{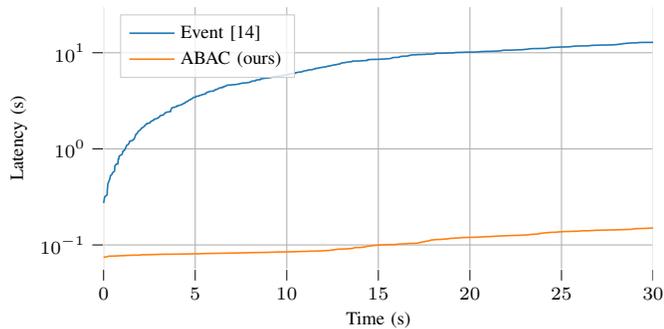}}
    \caption{This figure illustrates the importance of conflict resolution in the context of transaction commitment latency for ABAC and event-driven smart contracts, operating at a throughput of 50\,Hz.} 
    \label{fig:lat_hijack}
\end{figure}

To thoroughly evaluate the effectiveness of the proposed approach compared to the previous implementation~\cite{lei2023event}, we conducted an experiment that shed light on the importance of conflict resolution. In this experiment, we had two users attempting to send commands on the same topic at a frequency of 50Hz each, utilizing Attribute-Based Access Control (ABAC) and event-driven chaincodes. Both users were authorized members of the network; however, only one possessed the required attribute to publish on the topic. The experiment's results, depicted in Fig.~\ref{fig:throughput_hijack} for throughput and Fig.~\ref{fig:lat_hijack} for latencies, provide insightful observations.

Using the ABAC smart contract, the robot consistently received topic messages at a constant frequency of 50\,Hz from the user with the Turtlebot attribute, maintaining relatively stable latencies. This demonstrates the successful conflict resolution achieved by the ABAC approach. In contrast, when utilizing the event-driven smart contract, where both users sent 50\,Hz messages on the same topic, the robot encountered complications. Instead of the expected 100\,Hz message frequency, the robot received approximately 70\,Hz messages. Furthermore, the latency rapidly escalated, rendering robot control virtually impossible. This highlights the severe consequences of unresolved conflicts in the event-driven approach.

These findings emphasize the criticality of effective conflict resolution mechanisms, such as ABAC, in ensuring proper system behavior, maintaining desired frequencies, and mitigating latency escalation that can severely impact the robot's controllability.

Based on this work and previous studies~\cite{lei2023event}, it has been determined that Hyperledger Fabric has a maximum throughput capacity while maintaining the required latency levels for the system. It is essential that these performance characteristics align with the requirements of the specific multi-robot system. While Hyperledger Fabric provides ample opportunities, it is crucial to address the issue of unintended messages published on the fabric, which can disrupt the network. This unintended message artifact poses a risk to the integrity and reliability of the system. However, our proposed attribute-based access control mechanism offers a solution to mitigate such scenarios effectively.

With attributed-based access control, we show how Hyperledger Fabric brings security and performance to multi-robot systems. This mechanism allows for fine-grained control over access to resources, preventing unauthorized or unintended messages from being published on the fabric. Consequently, it strengthens the overall security posture of the system and mitigates the risks associated with unintended message artifacts. By adopting our proposed approach, multi-robot systems can leverage the benefits of Hyperledger Fabric's opportunities while maintaining the integrity and confidentiality of the network.

\section{Conclusion and Future Work}

In conclusion, this paper presents a customizable conflict resolution and attribute-based access control (ABAC) framework for multi-robot systems utilizing the Hyperledger Fabric network as an extension to the current SROS2 access control. The proposed framework offers a more flexible and configurable approach to conflict resolution and access control. This allows system administrators to define their own policies and rules to better fit their specific use case scenarios.

The experimental results demonstrate that the proposed framework effectively resolves conflicts and enforces access control policies. In addition, it does not compromise performance and scalability compared to the previous similar frameworks based on Hyperledger Fabric. Additionally, the proposed framework has the potential to support a wide range of multi-robot applications and enable the integration of various security and privacy mechanisms embedded in the chaincode.

In terms of future work, there are several avenues to explore further research and development in this area. One potential direction for future work would be to investigate the scalability and performance of the ABAC mechanism in large-scale multi-robot systems. Even though the ABAC mechanism presented in our paper provides a promising approach, we should explore how it scales as the number of robots and resources in the system grows. 

Additionally, further validation of the benefits of ABAC could be achieved through comparative studies with other access control mechanisms, like SROS2's role-based access control. It may be necessary to compare different access control mechanisms in multi-robot systems to see what they offer in terms of performance, scalability, and security properties. Such comparative studies could help shed light on the strengths and weaknesses of different access control mechanisms. They could also guide the development of more effective and efficient security solutions for multi-robot systems.

Finally, there is also the possibility of combining ABAC with other security mechanisms to provide a more comprehensive security solution. For example, the ABAC mechanism can be used with a distributed trust management system. This defines the level of trust between nodes based on ranking the data generated by robots.

\section*{Acknowledgment}

This research work is supported by the Academy of Finland's 
RoboMesh project (Grant No. 336061), and by the R3Swarms project funded by the Secure Systems Research Center (SSRC), Technology Innovation Institute (TII).

\bibliographystyle{unsrt}
\bibliography{bibliography}
\end{document}